\documentclass[conference]{IEEEtran}
\IEEEoverridecommandlockouts
\usepackage{cite}
\usepackage{amsmath,amssymb,amsfonts}
\usepackage{algorithmic}
\usepackage{graphicx}
\usepackage{textcomp}
\usepackage{xcolor}
\IEEEoverridecommandlockouts

\def\BibTeX{{\rm B\kern-.05em{\sc i\kern-.025em b}\kern-.08em
    T\kern-.1667em\lower.7ex\hbox{E}\kern-.125emX}}
\begin{document}

\title{Visual Task Modeling and Automated Planning\\for
Humanoid Robots in Education of Individuals\\with
Autism Spectrum Disorder}

\author{
\IEEEauthorblockN{Lada Kudláčková}
\IEEEauthorblockA{\textit{Department of Theoretical Computer Science and Mathematical Logic} \\
\textit{Charles University, Faculty of Mathematics and Physics}\\
Prague, Czech Republic \\
kudlackova@ktiml.mff.cuni.cz}
\thanks{\copyright~2026 IEEE. Personal use of this material is permitted. Permission from IEEE must be obtained for all other uses, in any current or future media, including reprinting/republishing this material for advertising or promotional purposes, creating new collective works, for resale or redistribution to servers or lists, or reuse of any copyrighted component of this work in other works.}%
\thanks{This is the accepted version of the paper. The final published version is: L. Kudláčková, ``Visual Task Modeling and Automated Planning for Humanoid Robots in Education of Individuals with Autism Spectrum Disorder,'' 2026 4th Cognitive Models and Artificial Intelligence Conference (AICCONF), Prague, Czech Republic, 2026, pp. 1--6, doi: 10.1109/AICCONF69182.2026.11600719. Available at https://doi.org/10.1109/AICCONF69182.2026.11600719}%
}
\maketitle

\begin{abstract}
In robot-assisted education, a robot must understand how to complete a task in order to effectively guide a student through it. We present a novel method that enables a semi-autonomous, camera-equipped robot to recognise the task goal and plan actions to achieve it. Our work focuses on Shoebox Tasks, an activity commonly used in the education of individuals with Autism Spectrum Disorder (ASD). These tasks are designed so that the goal can be inferred directly from their visual structure. We have developed a solver that uses a vision-based Large Language Model to convert photos of a Shoebox Task into a Planning Domain Definition Language (PDDL) representation and generate action plans for solving it. We have also implemented a robotic interface that enables a Nao robot to autonomously execute various Shoebox Tasks and assist a student by suggesting next steps or signalling detected failures through action demonstrations, voice commands, or gestures. With an adapted robotic interface, the solver can be integrated into a variety of educational robots. No programming expertise is required from teachers, and the solver operates exclusively on captured images as input. Research in special educational needs indicates that students with ASD can benefit substantially from robot-assisted teaching, and our system aims to strengthen these learning opportunities while also fostering interest in robotics among individuals with ASD.
\end{abstract}

\begin{IEEEkeywords}
visual task modeling, PDDL, Large Language and Vision Models, humanoid robots, robot-assisted education, Autism Spectrum Disorder.
\end{IEEEkeywords}

\section{Introduction}
\begin{figure}[!t]
\centering
\includegraphics[width=0.5\columnwidth]{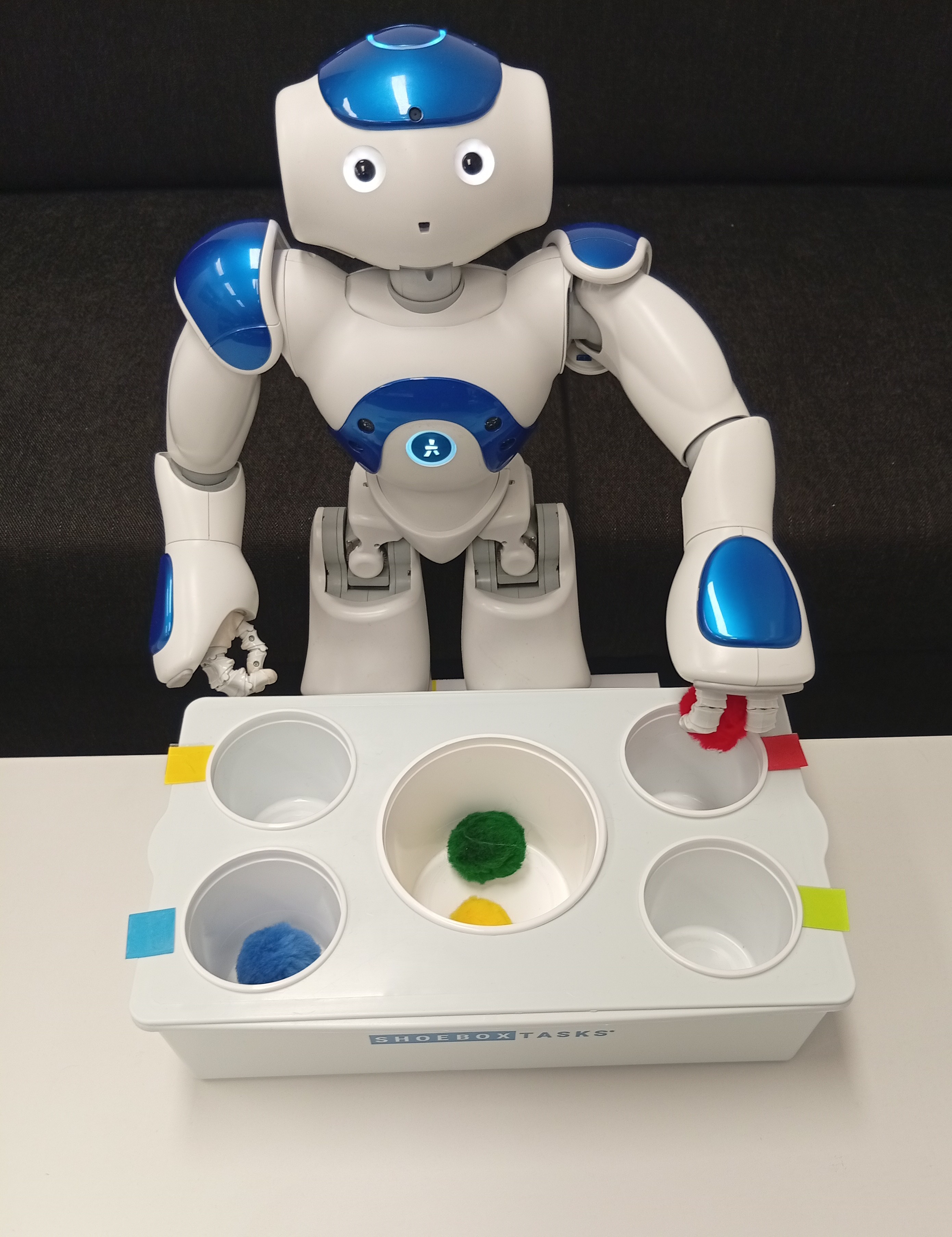}
\caption{The Nao robot performing the actions plan.}
\label{fig:nao_solving}
\end{figure}

We build a system that enables a robot to autonomously solve Shoebox Tasks and assist students in completing them. The automated task analysis relies entirely on visual input. We develop a task solver for Shoebox Tasks, an activity widely used in the education of individuals with Autism Spectrum Disorder (ASD). These tasks are intentionally highly visual, as learners with ASD often excel at visual learning but may struggle to follow spoken or textual instructions. Shoebox Tasks are therefore designed so that all information necessary to solve them can be inferred directly from their visual structure \cite{teacch}.\\
Our approach leverages this property by using Vision–Language Models (VLMs) to analyze images of the task, generate a corresponding Planning Domain Definition Language (PDDL) problem representation and action plan, and verify the plan execution through sequential image captures. We have implemented a simple robotic interface that allows a Nao humanoid robot to execute the generated plan autonomously or guide a student through the steps.\\
Robot-assisted education has the potential to enhance learning outcomes for people with ASD \cite{benef}. Learners may benefit from the robot’s predictability and consistency, and often respond positively to interaction with robots \cite{berk-smeekens}.\\
Our work builds on a recently introduced method observing that the strong visual structure of Shoebox Tasks—and similar educational activities—makes them well suited for VLM-based analysis \cite{kudlackova}. Initial experiments in that work used photos of the tasks with prompts and output examples as input to a VLM (e.g., ChatGPT-4o), producing a formatted analysis log and a PDDL problem file. This file was then passed to the Fast Downward planner to generate an action plan. After each executed action, a new image was captured, and the PDDL description and plan were updated iteratively until the task was completed.\\
We extend this idea by providing a simple, practical implementation of a task solver. The system incorporates a PDDL validator to enable automatic regeneration of output in cases of model false-positive refusals, and introduces image comparison technique to detect not just a single action but a sequence of executed actions.\\
Photos can be captured via a PC-connected camera or uploaded from a mobile phone through a web interface. Both task analysis and plan generation are executed directly by the VLM, which outputs a single PDDL problem file containing the action plan as a comment.\\
In the robotic setting, solving a Shoebox Task can be viewed as performing an unordered sequence of Pick-and-Place actions leading from an initial configuration to one of the allowed goal states. We have implemented a Nao robot interface enabling the robot to autonomously execute these actions or assist a student in solving selected types of Shoebox Tasks (Fig.~\ref{fig:nao_solving}).\\
We further observe that the proposed approach naturally supports the robot's learning by demonstration: users can teach the robot by manipulating objects while the system continuously updates its visual interpretation of the scene. This image-based control mechanism creates additional opportunities for robotics education, particularly for learners with ASD, as it enables the translation of physical demonstrations into PDDL-encoded actions. For example, each action can be represented by a card with a pictogram, allowing the robot to be controlled by presenting cards rather than through direct programming.\\

\subsection{Shoebox Tasks}
Shoebox Task is a visually structured educational activity consisting of a small number of movable items and clearly defined storage areas. The task is designed such that the student can infer both the required work and task completion solely from the visual structure, without textual or verbal instructions. Rather than a fixed task set, the Shoebox Task represents an approach for adapting general educational activities for students with ASD.\\
A simple example is a sorting task in which items placed on a table must be grouped by shape. In a Shoebox-style implementation, a box contains multiple smaller containers: one initial container holding the items and separate target containers marked with symbols corresponding to each shape.\\
The strong visual structure supports learners with ASD, who are predominantly visual learners, and also makes the task suitable for automated vision-based analysis.\\
ShoeboxTasks\textcopyright\ were originally developed in the 1990s by Ron Larsen as part of the TEACCH methodology \cite{teacch}, an educational framework widely used worldwide in the education of individuals with autism spectrum disorder (ASD). We had the privilege of collaborating with Ron Larsen to ensure that the proposed system closely aligns with the underlying educational principles of this methodology.\\

\subsection{Robot in the role of a teacher}
The robot acts as a teacher assisting the student during a Shoebox Task. In the standard setting, the student solves the task by manually moving items to their target containers, while a teacher provides assistance when needed by demonstrating actions, giving verbal instructions, pointing to relevant
items and containers, or alerting the student in case of an incorrect action.\\
The robot provides analogous, but simplified, feedback. It uses the solver to evaluate the validity of the current state and to generate an action plan. Based on the selected educational mode, the robot can alert the student about incorrect actions, demonstrate actions manually, or suggest the next action via voice commands or pointing gestures.\\
We implemented a solver prototype together with a basic robotic interface on the Nao robot to
demonstrate the feasibility of the proposed concept.\\

\subsection{Contributions}

The main contributions of this work are as follows:

\begin{itemize}
    \item Integration of vision-language reasoning with PDDL-based planning for Shoebox Tasks using a VLM (ChatGPT), without reliance on an external PDDL planner.
    
    \item End-to-end detection and validation of complete action sequences directly from images, enabling reasoning over both initial and intermediate Shoebox Tasks states.
    
    \item A simple interface for the Nao robot, allowing autonomous execution or assisted solving of Shoebox Tasks within the proposed framework.
\end{itemize}

\section{Method}

\subsection{System overview}
\begin{figure*}[htbp]
\centering
\includegraphics[width=0.6\textwidth]{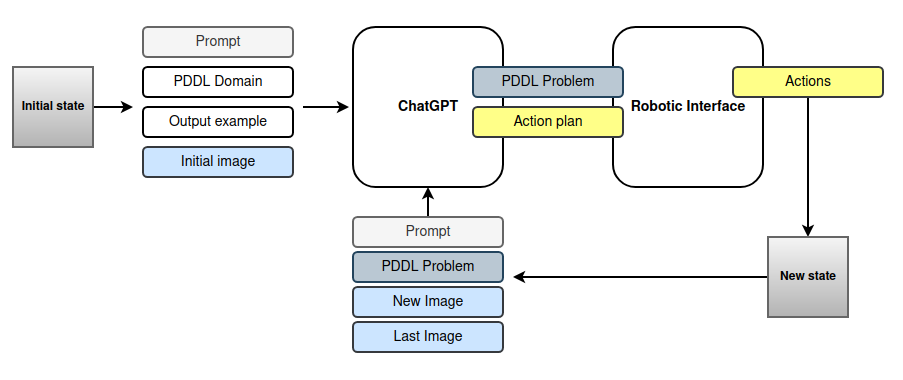}
\caption{The Perception–Analysis–Action Loop.}
\label{fig:loop}
\end{figure*}

Our system operates as a perception–analysis–action loop (Fig.~\ref{fig:loop}) implemented in Python and driven by a Vision–Language Model (VLM). Starting from an image of the initial Shoebox Task configuration and a general PDDL domain file, the system analyzes the visual structure and generates a corresponding PDDL problem instance under the given domain together with an action plan.\\
After one or more actions are performed, a new image is captured and provided to the system together with the previously generated PDDL file and the earlier image. The VLM infers which actions have been executed between states, removes them from the remaining plan, and updates the PDDL problem description accordingly. This loop continues until the task is completed or an invalid state is detected.\\
Actions may be executed autonomously by a robot, or performed by a student, with the robot providing guidance, error alerts, or corrective feedback.

\subsection{Vision–Language Analysis}
For the initial state, before any action is performed, the VLM is given:
\begin{itemize}
    \item an image of the start configuration,
    \item a general PDDL domain file,
    \item an example PDDL problem file,
    \item  a prompt containing analysis rules.
\end{itemize}

The VLM (ChatGPT-5) is instructed to detect objects in the image, identify which objects are movable and which are static containers, and infer the intended distribution of objects given the visual structure of the task.\\
The PDDL domain and the example problem file are designed to be general and applicable across a variety of Shoebox Tasks. The domain file defines three object types: \textbf{items}, \textbf{initial\_container}, and \textbf{targets}.  All movable items begin stacked inside a single initial container, placed side by side to avoid occlusion. All other static objects are considered targets, and the task is to determine how items should be distributed among them.\\
Detected objects are assigned types and deterministic names based on their visual features and numbered by their position in the image. Objects that share the same visual appearance are grouped and named using a descriptive label and an index (e.g. blue\_ball\_3). Unique objects receive unique descriptive identifiers. Clear naming is essential for correctly tracking objects across state updates.\\
The VLM produces a complete PDDL problem instance and an associated action plan consisting of an unordered set of valid actions. After one or more actions have been executed, a new image is captured and the VLM is given:
\begin{itemize}
    \item the latest PDDL problem file,
    \item the image of the latest (captured) configuration,
    \item the image of the current configuration.
\end{itemize}

The VLM, following the strict naming rules, re-identifies all objects, verifies whether the new state is valid, and determines which planned actions were executed. These actions are removed from the remaining plan and the corresponding predicates in the PDDL problem file are updated.\\
Once the plan is empty, the goal state has been reached. If the new state is invalid (e.g., an object disappears or is corrupted), the system reports a failure.

\subsection{PDDL Problem Generation}
The PDDL domain defines three object types: item, init\_container, and target.\\

The \textbf{predicates} are as follows:
\begin{itemize}
    \item (in\_container ?x – item),
    \item (in\_target ?x – item ?y – target),
    \item (free ?x – target).
\end{itemize}

\mbox{}\\
Two Pick\&Place \textbf{actions} are defined:
\begin{itemize}
    \item move\_to\_target, which transfers an item from the initial container to its designated target.
    \item move\_to\_container, which returns a misplaced item from a target back to the container.
\end{itemize}

The generated PDDL problem file contains the object list, initial state, and goal specification, followed by a comment section containing explanatory notes and the action plan. Object names follow the deterministic naming scheme, and the goal section specifies which item should be placed in which target.\\
Generated PDDL files are validated against the domain. When visual analysis is correct, the files are usually syntactically valid. However, VLMs occasionally produce false-positive refusals (e.g., “I’m sorry, I can’t assist with that”), which leads to PDDL validation failure. In such cases, the request is simply repeated.

\subsection{Robotic Interface}
We implemented a simple robotic interface for a Nao v5 humanoid robot to demonstrate how the solver can be used in robot-assisted education. Each suggested PDDL action is mapped to either a robot-executed manipulation or a robot-guided instruction, depending on the educational mode.\\
In manipulation mode, we restrict tasks to those involving objects that are easy for the robot to grasp (e.g. balls, cubes). To simplify manipulation and avoid solving complex inverse kinematics, items are placed at predefined positions within the initial container. In this mode, the robot performs only forward actions (moving items to targets), and recovery actions are not executed. After each step, a new image is captured.\\
Tasks requiring fine manipulation, such as using coins, are executed in voice-guidance mode, where the robot issues simple verbal instructions and gestures. Since the student may follow, ignore, or perform multiple actions at once, a new photo is captured on request to re-evaluate the state.\\
In this prototype implementation, the Nao’s onboard camera is not used. Instead, images are captured by a mobile phone or a PC-connected camera and uploaded through a web interface. This design allows the user (e.g., a teacher) to maintain strict control over the experimental process.\\
The image capture frequency is configurable and depends on the educational context. For example, when the robot demonstrates the task by executing the solution autonomously, a single initial image
is sufficient to compute the full action plan, which is then executed sequentially. However, to verify correct execution (e.g., detecting dropped items), images may be captured after each action.\\
When the student performs the task manually, images must be captured after each action (or at fixed time intervals) if the robot is expected to detect and alert incorrect actions. Alternatively, if the robot
provides assistance only upon student request, images may be captured after multiple actions.\\

\begin{figure*}[htbp]
\centering
\includegraphics[width=0.6\textwidth]{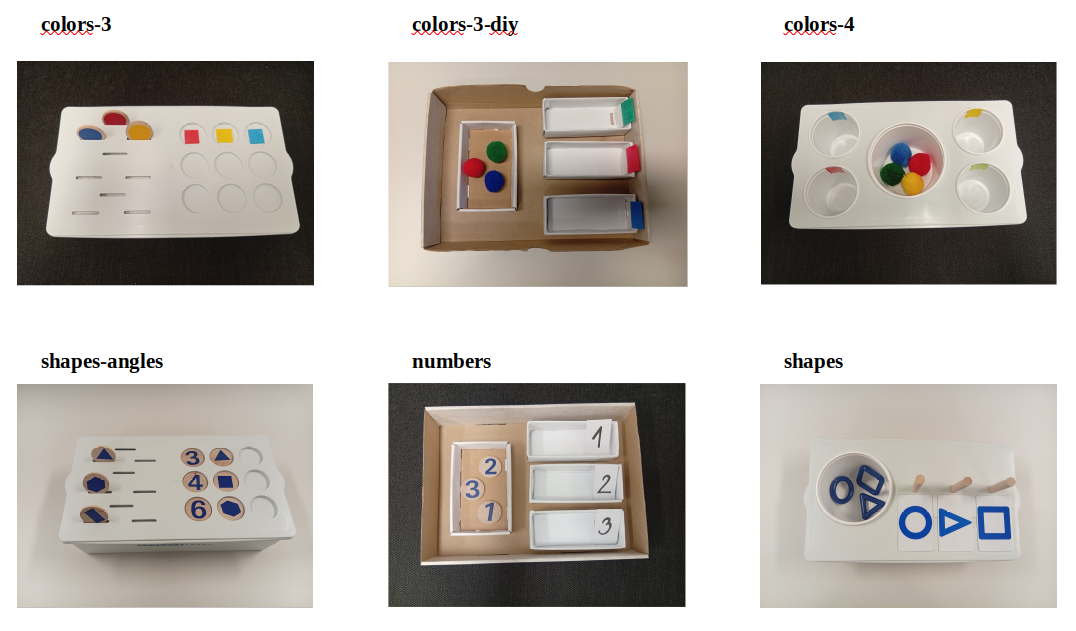}
\caption{Initial Model and Plan Experiment: Input images showing the initial configuration of six Shoebox Tasks.}
\label{fig:tasks}
\end{figure*}

\section{Experiments}

\subsection{Experimental Setup}

We conducted experiments exclusively on the solver component to evaluate the quality of the visual analysis, the correctness of the generated task model and its PDDL problem instances, and the validity of the produced action plan. All suggested actions were executed manually, without robot assistance.\\
Images were captured using a mobile phone and uploaded via a web interface. The solver was implemented using the OpenAI Python API, with experiments performed using ChatGPT-4o and ChatGPT-5.\\
During early development, we iteratively refined the prompt templates, the PDDL domain, and the example problem file using failure cases and counter-examples. After the prompts and domain stabilized, all reported experiments were executed using a single, fixed configuration.\\
The example problem file is intentionally kept simple and represents a task distinct from matching tasks, in which items are distributed randomly among the targets.\\

\subsection{Shoebox Tasks}

Shoebox Tasks encompass a broad range of pre-academic and pre-vocational activities. In this study, we focused on tasks involving a limited number of objects (up to nine), where each target corresponds to exactly one item. The selected tasks included one-to-one correspondence tasks with visually indistinguishable items, sorting tasks based on visual properties (e.g., color or shape), and sequencing tasks. Tasks primarily designed to train fine motor skills (e.g., requiring precise finger manipulation) were excluded.

Experiments were conducted using both original \textcopyright ShoeboxTasks and custom do-it-yourself (DIY) task boxes designed according to the same visual structuring principles. The items included colored balls and cubes placed into cups, wooden tokens with symbols matched to corresponding cutouts, and plastic shapes positioned on dowels with matching visual cues.

The initial configuration followed two layout variants: the container positioned on the left with targets on the right, and the container positioned centrally with four targets arranged in the corners.

\subsection{Initial Model and Plan}

We evaluated six Shoebox Tasks based on matching colors, shapes, and numbers, each involving three or four items. For each task, a photograph of the initial configuration was provided as input to the model (Fig.~\ref{fig:tasks}). The generated PDDL problem file was subsequently analyzed for correctness.

Each task was tested over ten iterations. For every trial, we measured the response time and evaluated the correctness of the generated PDDL output.

Experiments were conducted using two model versions, ChatGPT-4o and ChatGPT-5, accessed via the Python OpenAI API. Default parameter settings were used (i.e., no explicit temperature or reasoning configuration).
\\
\subsubsection{Correctness Evaluation Criteria}\mbox{}\\

\textbf{Semantic Errors:}
Incorrect representation of the problem or an invalid action plan. Examples include failure to detect all objects, incorrect object typing (e.g., confusing movable items with target locations), or incorrect goal inference.

\textbf{Inaccurate Names:}
Objects were correctly detected and typed, and the action plan was valid, but object names contained minor inaccuracies (e.g., \texttt{pentagon\_item} instead of \texttt{hexagon\_item}).

\textbf{Syntactic Errors:}
Invalid PDDL output that could not be parsed, typically due to un-commented natural language text or malformed syntax.

\textbf{Model Refusals:}
The model declined to generate a solution, returning a refusal message instead of analysis or PDDL content.

\begin{table}[t!]
\caption{Shoebox Task Performance Metrics: ChatGPT-4o}
\centering
\footnotesize
\setlength{\tabcolsep}{3pt} % reduce horizontal padding
\begin{tabular}{|c|r|c|c|c|c|}
\hline
\textbf{Shoebox Task} &
\textbf{\shortstack{Time \\ (s)}} &
\textbf{\shortstack{Semantic \\ Errors}} &
\textbf{\shortstack{Inaccurate \\ Names}} &
\textbf{\shortstack{Syntactic \\ Errors}} &
\textbf{\shortstack{Model \\ Refusals}} \\
\hline
colors-3           & 10.9 & 0 & 0  & 0 & 1 \\
colors-3-diy       & 8.9  & 0 & 0  & 0 & 0 \\
colors-4           & 12.0 & 0 & 0  & 0 & 0 \\
shapes-angles & 11.8 & 1 & 10 & 0 & 0 \\
numbers            & 11.2 & 0 & 0  & 0 & 0 \\
shapes             & 8.5  & 0 & 1  & 0 & 0 \\
\hline
\multicolumn{6}{l}{$^{\mathrm{a}}$All times are averages over ten trials, rounded to one decimal place.}
\end{tabular}
\label{tab:Shoebox1}
\end{table}

\begin{table}[t!]
\caption{Shoebox Task Performance Metrics: ChatGPT-5}
\centering
\footnotesize
\setlength{\tabcolsep}{3pt} % reduce horizontal padding
\begin{tabular}{|c|r|c|c|c|c|}
\hline
\textbf{Shoebox Task} &
\textbf{\shortstack{Time \\ (s)}} &
\textbf{\shortstack{Semantic \\ Errors}} &
\textbf{\shortstack{Inaccurate \\ Names}} &
\textbf{\shortstack{Syntactic \\ Errors}} &
\textbf{\shortstack{Model \\ Refusals}} \\
\hline
colors-3        & 75.8 & 0  & 0 & 0 & 0 \\
colors-3-diy    & 40.7 & 0  & 0 & 0 & 0 \\
colors-4        & 49.5 & 0  & 0 & 0 & 0 \\
shapes-angles   & 86.4 & 1  & 1 & 0 & 0 \\
numbers         & 42.9 & 0  & 0 & 0 & 0 \\
shapes          & 61.7 & 0  & 0 & 0 & 0 \\
\hline
\multicolumn{6}{l}{$^{\mathrm{a}}$All times are averages over ten trials, rounded to one decimal place.}
\end{tabular}
\label{tab:Shoebox2}
\end{table}

\subsubsection{Results}\mbox{}\\

Tables~\ref{tab:Shoebox1} and \ref{tab:Shoebox2} summarize the performance of ChatGPT-4o and ChatGPT-5 across the six evaluated Shoebox Tasks. The results reveal substantial differences in response time, while overall correctness remains comparable.

ChatGPT-4o achieved significantly shorter response times, ranging from 8.5\,s to 12.0\,s across all tasks. In contrast, ChatGPT-5 required considerably more time, with response times ranging from 40.7\,s to 86.4\,s. On average, ChatGPT-5 took approximately six times longer to produce a response than ChatGPT-4o.

In terms of correctness, both models demonstrated high reliability. ChatGPT-4o produced a single semantic error across all experiments. In the \textit{shapes-angles} task, all ten runs contained a naming inaccuracy, where the object and corresponding target marked with a hexagon were labeled as a pentagon. Across all six tasks, only one run resulted in an incorrect action plan and one run led to a model refusal. Furthermore, three out of six tasks were completed without any errors in all ten runs.

ChatGPT-5 produced one semantic error and one naming inaccuracy in the \textit{shapes-angles} task. No syntactic errors or model refusals were observed. Similarly, three out of six tasks were completed without any errors across all ten runs.

Overall, ChatGPT-4o demonstrated substantially faster response times while maintaining strong semantic and syntactic correctness. ChatGPT-5 exhibited slightly more stable behavior in terms of refusals and output consistency, albeit at the cost of significantly increased response latency.

\subsection{State Update and Action Detection}
The second experiment was conducted using a broader set of tasks and included analysis not only of the initial state but also of intermediate states during task execution, in order to demonstrate the effectiveness of the proposed framework. ChatGPT-5 was used for the analysis.
An image of the initial task configuration was captured and used to generate the PDDL problem and action plan. Actions were then executed manually by moving items between containers, after which a new image was captured and processed by the solver to update the task
state.\\
In educational setting, these actions would be performed either by the student or by the robot. When actions are performed manually by the student, the robot must infer which actions have already been executed in order to validate them and suggest appropriate next actions. This is
necessary because the action plan is an unordered set, as action execution order does not affect the final outcome.\\
Similarly, when the robot executes actions itself, the solver confirms the correct execution of the last action by removing the action from the action plan.\\
Results for the initial state and those for the subsequent state transitions were evaluated independently.\\

\subsubsection{Correctness Evaluation Criteria}\mbox{}\\

\noindent\textbf{Action Detection Accuracy:} The solver must correctly identify the actions that were performed.\\
\textbf{State Update Consistency:} The updated PDDL problem instance must accurately represent the new task state, both semantically and syntactically.\\
\textbf{Plan Update Correctness:} The remaining action plan must be updated to include only the actions still required to reach the goal state.\\
 \\
\subsubsection{Results}\mbox{}\\

\noindent\textbf{Small-Scale Tasks.} We obtained the most reliable performance on tasks containing up to four items and four targets (e.g. matching by color, shape, number, or symbol). All such tasks were solved successfully, with one notable exception: the task where three coins with triangle, square, and hexagon symbols must be matched to three targets marked by pair (number of angles, symbol).\\
In this task, the hexagon symbol was repeatedly misclassified as a pentagon, which led to incorrect naming (e.g. token\_pentagon, slot\_pentagon). Despite this naming error, the solver still inferred the correct goal configuration because the other two coins were classified correctly.\\
Apart from this case, the initial task model and action plan were consistently generated correctly, and state updates proceeded without errors. These tasks required no retries due to model refusal or invalid PDDL output.\\
\textbf{Medium-Scale Tasks.} For tasks containing five to nine items, the solver generally produced a correct initial model and action plan but we observed occasional errors. In one task, the initial container consisted of nine slots holding six coins, and the three empty slots were repeatedly misidentified as additional items. In all other cases, we eventually obtained a correct initial model and plan, typically with zero or one retry caused by model refusal.\\
Errors became more frequent during the state-update phase, especially for tasks containing nine items. Although correct results were usually obtained with one or two model refusals, several failures involved incorrect plan updates. In some instances, the executed actions were not identified correctly; in others, the solver detected the executed actions and updated the corresponding PDDL predicates, but failed to remove these actions from the plan.\\
Throughout all experiments, the generated PDDL files were syntactically valid whenever the model responded with PDDL content. The only invalid output occurred in cases of model refusal.\\

\section{Conclusion and Future Work}
We presented a practical, VLM-based task solver for Shoebox Tasks, capable of generating PDDL plans from visual input and identifying executed actions. Experimental results demonstrate reliable performance for tasks with up to four items and reasonable performance for medium-scale tasks. The robotic implementation, while simple, illustrates how the solver can be applied in the context of robot-assisted education.\\
We also observe that this approach naturally supports learning by demonstration and can be useful for teaching basic programming skills and fostering interest in robotics among individuals with ASD. Users can ``teach'' the robot simply by performing actions and saving the generated plans. Given that students with ASD often use visual schedules built from cards with images, we aim to extend the framework to map not only task images to actions but also activity cards to actions.\\
The visual structure leveraged in Shoebox Tasks extends beyond these simple activities to more complex tasks for individuals with ASD, such as interpreting visual diagrams on workshop tables during order assembly, and the framework can be extended to support these scenarios.\\
Future work will focus on increasing the robustness of the system to handle tasks with a larger number of objects and more complex configurations, such as occluded or hidden items in the initial state. Additionally, interaction with students requires the implementation of a more sophisticated communication layer for the robot, to maximize educational benefits and provide a safe, supportive, and motivating learning environment.\\

\section*{Acknowledgment}
This research was partially supported by the SVV project, project number 260 821, and the UNCE project.

\vspace{12pt}


\begin{thebibliography}{00}

\bibitem{teacch}
Centering on Children, ``TEACCH,'' 2018. [Online]. Available: https://www.Shoeboxtasks.com/teacch

\bibitem{benef}
U. Qidwai, S. B. A. Kashem, and O. Conor, ``Humanoid Robot as a Teacher’s Assistant: Helping Children with Autism to Learn Social and Academic Skills,'' J. Intell. Robot. Syst., vol. 98, pp. 759--770, 2020.

\bibitem{berk-smeekens}
I. van den Berk-Smeekens, M. van Dongen-Boomsma, M. W. P. De Korte, et al., ``Adherence and Acceptability of a Robot-Assisted Pivotal Response Treatment Protocol for Children with Autism Spectrum Disorder,'' Sci. Rep., vol. 10, p. 8110, 2020.

\bibitem{kudlackova}
L. Kudláčková, ``Vision-Based Task Analysis and Planning for Robot-Assisted Education of Individuals with Autism Spectrum Disorder,'' in \textit{Robotics in Education. RiE 2025}, R. Balogh, D. Obdržálek, and N. Fachantidis, Eds., Lecture Notes in Networks and Systems, vol. 1544. Cham: Springer, 2025.

\end{thebibliography}
\end{document}